\documentclass[journal]{IEEEtran}

\ifCLASSINFOpdf
\else
   \usepackage[dvips]{graphicx}
\fi
\usepackage{url}

\usepackage{graphicx}
\usepackage{bm}
\usepackage{algorithm}
\usepackage{algorithmic}
\usepackage{multirow}
\usepackage{color} 
\usepackage[pagebackref=true,breaklinks=true,letterpaper=true,colorlinks,bookmarks=false]{hyperref}

\usepackage{graphicx}
\usepackage{float} 
\usepackage{subcaption}
\usepackage{amsthm}
\usepackage{tikz}
\usepackage{comment}
\usepackage{amsmath,amssymb} 
\usepackage{color}
\usepackage{amsmath}
\usepackage{amssymb}
\usepackage[accsupp]{axessibility}  
\usepackage{multirow}
\usepackage{amsfonts}
\usepackage{graphicx}
\usepackage{amsmath}
\usepackage{amsthm}
\usepackage{booktabs}
\usepackage{bm}
\usepackage{algorithm}
\usepackage{algorithmic}
\usepackage{marvosym}

\begin{document}

\title{NeuroCLIP: Neuromorphic Data Understanding by CLIP and SNN}

\author{Yufei Guo, Yuanpei Chen, and Zhe Ma
\thanks{This paragraph of the first footnote will contain the date on which you submitted your paper for review. It will also contain support information, including sponsor and financial support acknowledgment. This work was supported in part by the National Natural Science Foundation of China under Grant No.12202412 and No.12202413.}
\thanks{Yufei Guo and Yuanpei Chen contribute equally to the paper (e-mail: yfguo@pku.edu.cn, rop477@163.com). Zhe Ma is the corresponding author (e-mail: mazhe\_thu@163.com).
}
\thanks{All authors are with the Intelligent Science \& Technology Academy of CASIC, China.}}

\markboth{Journal of \LaTeX\ Class Files, Vol. 14, No. 8, August 2015}
{Shell \MakeLowercase{\textit{et al.}}: Bare Demo of IEEEtran.cls for IEEE Journals}
\maketitle

\begin{abstract}
Recently, the neuromorphic vision sensor has received more and more interest.
However, the neuromorphic data consists of asynchronous event spikes, which makes it difficult to construct a big benchmark to train a power general neural network model, thus limiting the neuromorphic data understanding for ``unseen” objects by deep learning. 
While for the frame image, since the training data can be obtained easily, the zero-shot and few-shot learning for ``unseen” task via the large Contrastive Vision-Language Pre-training (CLIP) model, which is pre-trained by large-scale image-text pairs in 2D, have shown inspirational performance. 
We wonder whether the CLIP could be transferred to neuromorphic data recognition to handle the ``unseen” problem.
To this end,  we materialize this idea with NeuroCLIP in the paper.
The NeuroCLIP consists of 2D CLIP and two specially designed modules for neuromorphic data understanding.
First, an event-frame module that could convert the event spikes to the sequential frame image with a simple discrimination strategy.
Second, an inter-timestep adapter, which is a simple fine-tuned adapter based on a spiking neural network (SNN) for the sequential features coming from the visual encoder of CLIP to improve the few-shot performance.
Various experiments on neuromorphic datasets including N-MNIST, CIFAR10-DVS, and ES-ImageNet demonstrate the effectiveness of NeuroCLIP.
Our code is open-sourced at \href{https://github.com/yfguo91/NeuroCLIP.git}{NeuroCLIP}.
\end{abstract}


\IEEEpeerreviewmaketitle

\section{Introduction}
\label{sec:intro}

\IEEEPARstart{R}{ecently}, the neuromorphic vision sensor, or ``event camera'', such as~\cite{brandli2014240,posch2010qvga,gallego2017event,gallego2020event}, has received growing attention. It works asynchronously on the pixel level, that each pixel measures the incoming light intensity independently and fires an event when the absolute intensity change exceeds a certain threshold. Due to this unique information processing paradigm, it enjoys many significant advantages compared to the conventional frame-based camera, like high dynamic range (more than 120 dB), low latency (about 1 \textit{us}), and sparse nature of the events. Thus far, the neuromorphic vision sensor has been widely adopted in various applications, such as feature detection and tracking~\cite{Conradt2009,Wang2016}, optical flow estimation~\cite{Lagorce2014Event, 2019Event}, 3D reconstruction monocular and stereo~\cite{2011Event,2012Live}, pose estimation and SLAM~\cite{2016Real,Henri2017EVO}, image reconstruction~\cite{2014Real,2019Bringing}, motion segmentation~\cite{2018Simultaneous,2019Event},  and neuromorphic control~\cite{2016EventControl,Heemels2012An}.

However, the neuromorphic data format is not natural and very different from the frame data, thus it is not easy to construct a big neuromorphic data benchmark, which limits the neuromorphic data understanding by deep learning, especially to recognize these ``unseen” objects, which are not adopted as training samples but need to be recognized in the inference. This task is also named zero-shot recognition.
While the frame data is widely used in our lives and easy to collect, thus the ``unseen” object recognition problem has been dramatically mitigated in 2D vision by Contrastive Vision-Language Pre-training (CLIP)~\cite{radford2021learning}, a general model which is pre-trained by large-scale image-text pairs. 
It is shown that the performance of CLIP even has surpassed the networks well-trained on some full datasets~\cite{zhou2022cocoop,zhou2022coop}. 
Considering the success of the CLIP for ``unseen” object understanding, some works have applied it to the other modal data, \textit{e.g.}, point 
cloud~\cite{zhang2021pointclip,hess2023lidarclip}, satellite imagery~\cite{liu2023remoteclip}, mesh~\cite{khalid2022clipmesh}, and so on. Naturally, a question arises: could the CLIP also be generated for neuromorphic data understanding to realize the zero-shot task for ``unseen” event objects? If so, this would provide a promising router to deal with the ``unseen” problem for neuromorphic data.

\begin{figure*}[t]
	\centering
	\includegraphics[width=0.95\textwidth]{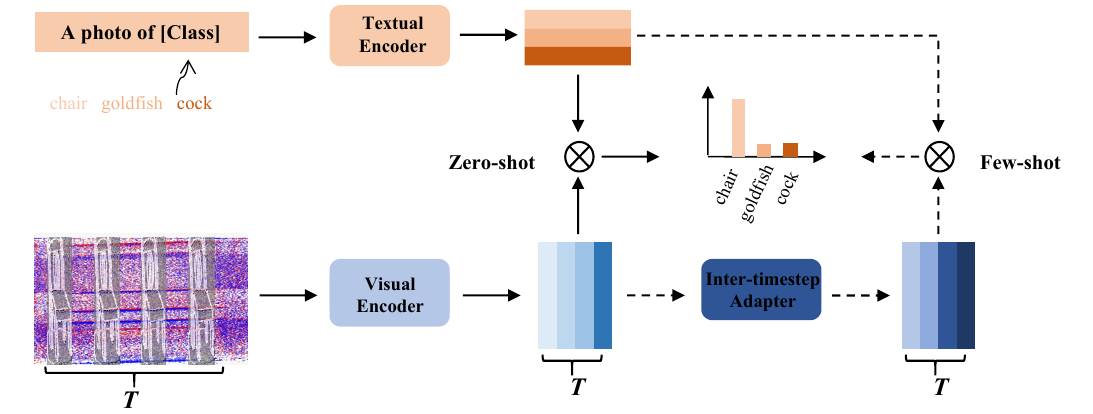} 
	\caption{The framework of NeuroCLIP. The NeuroCLIP first projects the neuromorphic data flow to multi-timestep frames and then conducts neuromorphic data recognition via CLIP pre-trained in 2D. For better few-shot classification, it also provides a learnable inter-timestep adapter based on a spiking neural network.}
	\label{titlefig1}
\end{figure*}

Thereby, in the paper, we realize the idea with the NeuroCLIP, which uses the CLIP’s 2D pre-trained knowledge to understand the neuromorphic data. Two special modules are proposed to assist the cross-modality transfer of the CLIP. First, an event-frame module is proposed to convert the neuromorphic data flow to $T$ sequential images. For the real-time application, we project the neuromorphic data flow to multi-timestep frames just in a simple manner that divides the event sequence into $T$ time intervals and generates a frame in each interval by discriminating different events over time. Then with the converted images, we can generate $T$ outputs using CLIP. The final result is acquired by their weighted summation as in~\cite{zhang2021pointclip}. 
Second, to further improve the performance of neuroCLIP, a learnable inter-timestep adapter to fine-tune the model is also proposed. 
Even though the adapter can be fulfilled by a CNN-based network, the temporal information behind the sequential event images cannot be effectively used. In view of this, we realize the adapter based on a spiking neural network, which has been proven can efficiently consider and utilize both spatial and temporal information of event images~\cite{guo2023direct}. 
The framework of our method is illustrated in Fig.~\ref{titlefig1}. 
Our contributions are as follows:
\begin{itemize}
	
\item We propose the neuroCLIP to transfer the CLIP's pre-trained knowledge to handle the ``unseen” problem for neuromorphic data, which as far as we know, is one of the few works focusing on transfer learning in this field. We believe that our proposed method would lay a new foundation to understand the ``unseen” object of the neuromorphic data for future research.

\item We also propose a learnable inter-timestep adapter based on a spiking neural network to fine-tune the neuroCLIP, which could efficiently utilize both spatial and temporal information of event images in few-shot settings to improve the performance of the neuroCLIP further.

\item Extensive experiment results on N-MNIST, CIFAR10-DVS, and ES-ImageNet show that the NeuroCLIP can deal well with the zero-shot or few-shot problems for neuromorphic data. 

\end{itemize}

\section{Methodology}

In this section, we first simply revisit the CLIP for 2D vision understanding. Then we introduce our NeuroCLIP for zero-shot recognition of the neuromorphic data. Finally, we present an inter-timestep adapter for better few-shot performance.

\subsection{A revisit of Contrastive Vision-Language Pre-training}

CLIP is a model that could jointly understand the image and text content. It is pre-trained on a large dataset consisting of 400 million image-text pairs collected from the internet, thus having a powerful ability to extract transferable visual features and the correlation to natural language description, known as the prompt. 
Based on the diverse textual understanding ability, the CLIP has shown promising performance in open-vocabulary recognition.
In specific, the model consists of two encoders, a text encoder $\mathcal{F}_T$ and an image encoder $\mathcal{F}_V$, both of which yield a single feature vector describing their input.
For an “unseen” picture from a dataset of $K$ classes, one could construct $K$ text prompts, \textit{e.g.}, ``a photo of a [class name]'',  CLIP will generate $K$ textual features of category texts with $C$-dimensions by the textual encoder denoted as ${\bm W}_t \in \mathbb{R}^{K \times C}$ and a $C$-dimensional image feature by the visual encoder denoted as ${\bm f}_v \in \mathbb{R}^{1 \times C}$. 
Then the classification probability $ {\bm p} \in \mathbb{R}^{1 \times C} $ can be computed as, 
\begin{equation}
    {\bm p} = {\rm SoftMax}({\bm W}_t {\bm f}_v),
\end{equation}
where ${\rm SoftMax}(\cdot)$ and ${\bm p}$ denote the softmax function and predicted probability for $K$ classes.
In theory, any object encountered in the 400 million text-image pairs could be recognized with this approach.

Due to the powerful ability of the CLIP, many works have tried to transfer the model to other domains. For example, PointCLIP~\cite{zhang2021pointclip} generalized the CLIP to the 3D zero-shot recognition on point cloud without any 3D training. LidarCLIP~\cite{hess2023lidarclip} could encodes lidar data into
an existing text-image embedding space. It is trained using image-lidar pairs assisted by the pre-trained CLIP, connecting lidar to both text and images. 
CLIP-Mesh~\cite{khalid2022clipmesh} presented a technique for the zero-shot generation of a 3D model using only a target text prompt based on the CLIP.
Wav2CLIP~\cite{WuWav2CLIP} proposed a robust audio representation learning method by distilling from CLIP.
In the paper, we focus on adapting the CLIP for the neuromorphic data.

\subsection{Zero-shot Classification by NeuroCLIP}
\textbf{Constructing frames from the neuromorphic data.}
Neuromorphic data is a set of asynchronous event spikes, which greatly differ from grid-based 2D images, thus cannot be used for CLIP directly.
Given a neuromorphic data flow with pixel grid size $M \times N$, it can be denoted as a sequence as
\begin{equation}
    \varepsilon = \{e_i\}_{i=1}^I, \ {\rm with} \ e_i = (x_i,y_i,t_i,p_i),
\end{equation}
where $x_i \in [1, \cdots, M]$ and $y_i \in [1, \cdots, N]$ denote the coordinates of the pixel which generates the event, $t_i \geq 0$ denotes 
the timestamp at which the event is generated,  
$p_i \in \{-1, 1\}$
denotes the polarity of the generated event, meaning OFF and ON respectively, and $I$ is the number of events. 
To realize the real-time neuromorphic data to frame conversion, we present a spatio-temporal representation for events here by making a balance between computational complexity and information integrity.
Given a pre-setting number of events, \textit{e.g.}, $T$, we first fairly divide the event sequence into $T$ time intervals. In each interval, an event frame can be generated by determining if there have been any events over time through each pixel. Specifically, we simply set the background as 127, then if there has been an ON event on any pixel, the pixel will be set as 255, otherwise 0 for the OFF event following~\cite{sironi2018hats}, given by
\begin{equation}
    \begin{split}
    \bm{I}_{t,x,y} = 
    \begin{cases}
        255 & \text{if there is a ON event in $t$-th interval,} \\
        0 & \text{if there is a OFF event in $t$-th interval,}\\
        127 & \text{otherwise,}
    \end{cases} \\
 \label{eq_fire}
    \end{split}
\end{equation}
where ${I}_{t,x,y}$ denotes the intensity of the pixel ${x,y}$ of the $t$-th image. Based on the number of time intervals, the raw event data can be converted to $T$ event frames, known as timesteps.

\noindent \textbf{Zero-shot understanding by NeuroCLIP.}
Since the raw spike data will be split into $T$ frames, we can get $T$ image features based on the visual encoder of the CLIP, denoted as $f_v = \{f_{vi}\}$, for $i = 1, 2, \cdots, T$. 
For the textual features, we construct the predefined template with these category names first. 
Then we can get the text embeddings ${\bm W}_t \in \mathbb{R}^{K \times C}$ from the textual encoder of the CLIP.
On top of that, the classification probability for these frames for the event sequence can be computed as 
\begin{equation}
    p = {\rm SoftMax}(\sum \alpha_i W_t f_{vi}),\ {\rm for} \  i = 1, 2, \cdots, T,
\end{equation}
where $\alpha_i$ is a hyper-parameter weighing the importance of frames. 
We set $\alpha_i = \frac{1}{T}$ in the paper.
So far, we have completed the zero-shot classification of the neuromorphic data without any extra training.
It provides a new paradigm to understand the ``unseen” object of the neuromorphic data.

\subsection{Few-shot Classification by NeuroCLIP}
Although applying the CLIP to the neuromorphic data can achieve relatively comparable zero-shot classification, there is also some work to further improve its performance on other data with few-shot setting~\cite{zhang2021pointclip,hess2023lidarclip}. Here we also consider the common scenario where a few samples are available
in each category for model training. 
With limited data but enormous parameters, it is not easy to fine-tune the whole model as it will easily result in over-fitting. 
Thus, we propose a light adapter after the visual encoder after the CLIP only fine-tuning the visual feature while keeping the CLIP's parameters frozen.
To better utilize the temporal information of the neuromorphic data, we designed the inter-timestep adapter based on a spiking neural network.

\textbf{Inter-timestep Adapter for NeuroCLIP.}
It is shown that the spiking neural network is suitable for handling the temporal neuromorphic data~\cite{guo2023direct,guo2023rmploss,guo2023membrane,guo2022imloss,Guo_2022_CVPR,guo2023ternary}. 
There are many excellent works dealing with combining SNNs and neuromorphic data. For example,
in~\cite{kosta2022adaptive}, an optical flow estimation method was presented based on the SNN and event data. 
In SuperFast~\cite{gao2022superfast}, an event-enhanced high-speed video frame interpolation method leveraging an SNN and a neuromorphic camera is presented. 
E-SAI~\cite{yu2022learning} provided a novel synthetic aperture imaging method, which can see through dense occlusions and extreme lighting conditions from event data, based on the combination of the SNN and neuromorphic camera.
Some works apply the combination of the SNN and the neuromorphic data in autonomous driving.
In~\cite{cordone2022object}, a fast and efficient automotive object detection method with SNNs on automotive neuromorphic data was proposed. 
In~\cite{Renner2020Event}, an attention mechanism based on a recurrent SNN, was proposed for object tracking. there are also some optimizing methods~\cite{KIM2021686} and augmentation methods~\cite{Yuhang2022Neuromorphic} proposed to improve the performance of the SNN on neuromorphic data.
All these works have demonstrated the superior advantage of SNN in handling neuromorphic data. 
Here we also use the SNN to construct our inter-timestep adapter. 
We use the widely used Leaky-Integrate-and-Fire (LIF) neuron model~\cite{nahmias2013leaky} to implement the SNN. This kind of spiking model uses the rich membrane potential to realize the unique spatial-temporal dynamic.
In LIF, the membrane potential is updated by
\begin{equation}
    \bm{u}^{(t+1), \text{pre}} = \tau\bm{u}^{(t)} + \bm{c}^{(t+1)}, \text{where } \bm{c}^{(t+1)} = \mathbf{W} \bm{x}^{(t+1)}, \label{eq_lif}
\end{equation}
where $\bm{u}$ is the membrane potential and $\bm{u}^{(t+1), \text{pre}}$ is the updated membrane potential at time step $t+1$, $\bm{c}^{(t+1)}$ is the pre-synaptic input at time step $t+1$, which is charged by weight-summed input spikes $\bm{x}^{(t+1)}$, and $\tau$ is a constant within $(0, 1)$, which controls the leakage of the membrane potential.
The LIF spiking neuron will fire a spike, when the updated membrane potential $\bm{u}^{(t+1), \text{pre}}$ is up to the firing threshold $V_{\rm th}$, as bellow,
\begin{equation}
    \begin{split}
    \bm{o}^{(t+1)} = 
    \begin{cases}
        1 & \text{if } \bm{u}^{(t+1), \text{pre}} > V_{\rm th} \\
        0 & \text{otherwise}
    \end{cases}, \\
    \bm{u}^{(t+1)} = \bm{u}^{(t+1), \text{pre}}\cdot(1 - \bm{o}^{(t+1)}). \label{eq_fire}
    \end{split}
\end{equation}
After firing, the spike output $\bm{o}^{(t+1)}$ at time step $t+1$ will be transmitted to the next layer and become its input. At the same time, the updated membrane potential will be reset to zero and becomes $\bm{u}^{(t+1)}$ to join the neuron processing at the next time step.

\begin{figure}[t]
	\centering
	\includegraphics[width=0.48\textwidth]{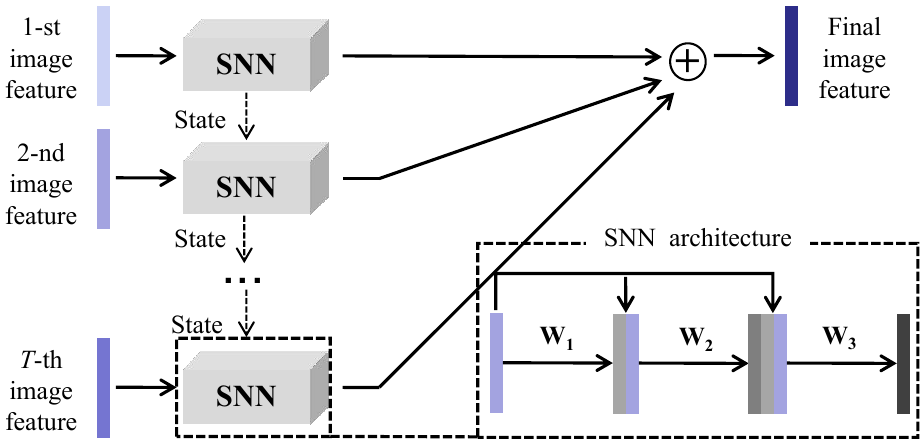} 
	\caption{The detailed structure of the proposed inter-time Adapter.}
	\label{titlefig}
\end{figure}

As illustrated in Fig.~\ref{titlefig}, our SNN inter-timestep adapter
consists of a three-layer structure composed of LIF neurons. the $f_{vi}$ coming from the CLIP will be fed into the spiking inter-timestep adapter sequentially over time, and the membrane potential of spiking
neurons of the SNN adapter will update between time intervals. The unique spatial-temporal dynamic membrane potentials will capture the temporal information between these features $f_{vi}$ and generate the new temporal-related features $\hat{f}_{vi}$. As in other works, we also sum these generated features as a final feature $\hat{f}_{v}^{\rm final}$ to represent the neuromorphic data flow. The process of the inter-timestep adapter can be denoted as 
\begin{equation}
    \begin{split}
    \hat{f}_{v}^{\rm final} = \frac{1}{T}\sum({\rm ReLU} (W_3{\rm Concate}(f,{\rm LIF}(W_1f_{vi}), f_{vi})),\\
    {\rm where} \quad f = {\rm LIF}(W_2{\rm Concate}({\rm LIF}(W_1f_{vi}),f_{vi})).
    \end{split}
\end{equation}
In the above equation, $W_1$, $W_2$, and $W_3$ are the weights for the three fully-connected layers. To increase accuracy, we use {\rm ReLU} in the last layer.
Then the classification probability for the few-shot setting will be updated as 
\begin{equation}
    p = {\rm SoftMax}(W_t \hat{f}_{v}^{\rm final}).
\end{equation}
\section{Experiments}
In this section, We evaluated our method using ResNet50, ResNet101, ViT-B/16, and ViT-B/32 as backbones on three well-known datasets: N-MNIST~\cite{orchard2015converting}, CIFAR10-DVS~\cite{2017CIFAR10}, and ES-ImageNet~\cite{ES-ImageNet}. N-MNIST dataset is a spiking version of the original frame-based MNIST dataset, consisting of the same 60,000 training and 10,000 testing samples. CIFAR10-DVS contains 10,000 128×128 images converted from CIFAR10. ES-ImageNet is a large-scale neuromorphic dataset converted from ImageNet with about 1,300,000 samples. As in most other works, we also convert the neuromorphic data to 10, 10, and 8 frame images for N-MNIST, CIFAR10-DVS, and ES-ImageNet respectively. The accuracy results are shown in Tab.~\ref {result}.   

\subsection{Zero-shot Classification}
In the zero-shot classification task, we directly converted the neuromorphic data to the frame image and used the pre-trained CLIP in 2D to evaluate the full test set for all datasets. 
To align with CLIP’s settings, we upsampled all converted images to (224, 224).
It can be seen from Tab.~\ref {result} that without any training, NeuroCLIP is able to achieve considerable performance, even 46.19\% on N-MNIST, which demonstrates the effective knowledge transfer from the image dataset to the neuromorphic dataset. 
We also find that the backbone choice is very important for NeuroCLIP, that accuracy on N-MNIST is 46.19\% for the ResNet-101, while only 21.65\% for the ViT/32. The same phenomenon also occurs in other datasets. 

We also conducted an ablation study with different prompt designs on CIFAR10-DVS~\cite{2017CIFAR10} using ResNet50 as the backbone. The results are shown in Tab.~\ref {ablation}.
We observe that without any extra description, the ``[CLASS].” achieves 15.00\%. We then insert the words ``a photo of", ``an image of", and ``a picture of" into it respectively. The accuracy will be improved by 3.80\%, 1.90\%, and 4.70\% respectively. Finally, we designed the prompt elaborately as ``a low quality stick figure of the object [CLASS].", the NeuroCLIP will reach the best-performing 21.40\%, demonstrating the importance of prompt choices.

\begin{table}[]	
	\begin{center}	
	\caption{The accuracy results for NeuroCLIP.}	
	\label{result}
  \resizebox{0.48\textwidth}{!}{
	\begin{tabular}{llllll}
		\toprule
  \midrule 
  \multirow{2}{*}{\bf Dataset}   & \multirow{2}{*}{\bf Type}  & \multicolumn{4}{c}{\bf Model}\\
\cmidrule{3-6}
        &         &      RN50   &    RN101  &   ViT/16  &  ViT/32 \\
\midrule     
		\multirow{3}{*}{N-MNIST} & Zero-shot & 44.35\%  & 46.19\% & 39.02\% & 21.65\%\\
    & 8-shot & 89.37\%  & 87.56\% & 90.01\% & 84.32\%\\
  & 16-shot & 90.40\%  & 89.12\% & 91.51\% & 85.64\%\\
\midrule   
		\multirow{3}{*}{CIFAR10-DVS} & Zero-shot & 21.40\%  & 18.40\% & 18.20\% & 17.20\%\\
    & 8-shot & 42.30\%  & 40.60\% & 39.70\% & 37.10\%\\
  & 16-shot & 44.40\%  & 43.80\% & 41.40\% & 39.30\%\\
\midrule 
		\multirow{3}{*}{ES-ImageNet} & Zero-shot & 5.26\%  & 8.58\% & 5.49\% & 3.54\%\\
    & 8-shot & 8.31\%  & 10.23\% & 8.22\% & 4.83\%\\
  & 16-shot & 12.49\%  & 14.72\% & 11.23\% & 7.67\%\\
\midrule 
		\bottomrule 	
	\end{tabular}}	
 \end{center}
 {$^\star$ RNn and ViT-B/n denote ResNet-n and vision transformer with n × n patch embeddings.}
\end{table}

\begin{table}[]	
	\begin{center}	
	\caption{Performance of NeuroCLIP with different prompt designs
on CIFAR10-DVS.}	
	\label{ablation}
  \resizebox{0.48\textwidth}{!}{
	\begin{tabular}{lc}
		\toprule
  \midrule 
  {\bf Prompt}  & {\bf Accuracy}\\
\midrule     
	``[CLASS]" & 15.00\%\\	
  ``a photo of [CLASS]" & 18.80\%\\
        ``an image of [CLASS]" & 16.90\%\\
        ``a picture of [CLASS]" & 19.70\%\\
        ``a low quality stick figure of the object [CLASS]." & 21.40\%\\
		\bottomrule 	
	\end{tabular}}	
 \end{center}
\end{table}

\subsection{Few-shot Classification}
Furthermore, we also evaluate the 8-shot and 16-shot recognition ability of NeuroCLIP on these datasets. 
For the inter-time adapter, we construct a residual-style multi-layer SNN consisting of three linear layers, as shown in Fig.~\ref{titlefig}.
With only 8 shots per class, the performance will be improved hugely compared with the setting of no training, even boosted to 90.01\% on N-MNIST. This shows the effectiveness of the inter-timestep adapter and the NeuroCLIP. 
When there are 16 samples per category, the accuracy can be improved further. Note that the improvement on the ES-ImageNet is more obvious, which means that these large datasets need more training samples to squeeze the few-shot ability of NeuroCLIP.

\section{Conclusion}
In the paper, we proposed NeuroCLIP for neuromorphic data understanding, which as far as we know, is one of the few works solving the “unseen” tasks in this field.
The proposed NeuroCLIP can conduct zero-shot recognition on neuromorphic data efficiently without any extra training. 
Additionally, we also propose a lightweight inter-time adapter based on a spiking neural network that can efficiently utilize both spatial and temporal information of event images to fine tune the CLIP to improve the performance of NeuroCLIP further under few-shot settings.
We have conducted extensive experiments on N-MNIST, CIFAR10-DVS, and ES-ImageNet with different backbones.
The results show that our NeuroCLIP can handle the zero-shot and few-shot problems of neuromorphic data well.
We believe the NeuroCLIP would provide a new way to understand the ``unseen'' neuromorphic object for future research.


\bibliographystyle{IEEEbib}
\bibliography{strings,refs}

\end{document}